\documentclass[preprint]{article} 

\usepackage[nonatbib]{neurips_2025}
\usepackage{amsmath}
\usepackage[utf8]{inputenc} 
\usepackage[T1]{fontenc}    
\usepackage{hyperref}       
\usepackage{url}            
\usepackage{booktabs}       
\usepackage{graphicx}       
\usepackage{amsfonts}       
\usepackage{nicefrac}       
\usepackage{microtype}      
\usepackage{xcolor}         
\usepackage{multirow}
\usepackage[
backend=biber,
maxcitenames=1,
style=authoryear,
sorting=ynt
]{biblatex}
\title{Cross-Attention Speculative Decoding}

\author{%
  Wei Zhong \\ 
  \texttt{wei.zhong@lge.com} \\
  \And
   Manasa Bharadwaj \\
  \texttt{manasa.bharadwaj@lge.com} \\
  \And
  Yixiao Wang \\
  \texttt{yixiao.wang@lge.com} \\
  \And
  Yipeng Ji \\
  \texttt{yipeng.ji@lge.com} \\
  \And
  Chul Lee \\
  \texttt{clee.lee@lge.com}
}

\begin{document}

\maketitle

\begin{abstract}
Speculative decoding (SD) is a widely adopted approach for accelerating inference in large language models (LLMs), particularly when the draft and target models are well aligned. However, state-of-the-art SD methods typically rely on tightly coupled, self-attention-based Transformer decoders, often augmented with auxiliary pooling or fusion layers. This coupling makes them increasingly complex and harder to generalize across different models.
We present Budget EAGLE (Beagle), the first, to our knowledge, cross-attention-based Transformer decoder SD model that achieves performance on par with leading self-attention SD models (EAGLE-v2) while eliminating the need for pooling or auxiliary components, simplifying the architecture, improving training efficiency, and maintaining stable memory usage during training-time simulation. To enable effective training of this novel architecture, we propose Two-Stage Block-Attention Training, a new method that achieves training stability and convergence efficiency in block-level attention scenarios.
Extensive experiments across multiple LLMs and datasets show that Beagle achieves competitive inference speedups and higher training efficiency than EAGLE-v2, offering a strong alternative for architectures in speculative decoding.
\end{abstract}

\section{Introduction}
Speculative decoding (SD)~\parencite{stern2018blockwise,sun2021early_gramar_err,xia2022firstSD,leviathan2023sampling,chen2023sampling,xia2024survey} is an effective method for accelerating inference in large language models (LLMs), where a lightweight draft model proposes the next $n$ tokens in advance, reducing the need for multiple target model invocations. The adoption of SD has been growing in industry due to its ability to deliver lossless latency improvements in both greedy and sampling-based decoding ~\parencite{leviathan2023sampling,chen2023sampling}, while also improving utilization of otherwise idle compute during memory-bound decoding phases.

Implementing SD efficiently typically requires replacing generic, often misaligned draft models with dedicated ones that are co-trained alongside the target model to match its output distribution better. As a result, the design and integration of SD models pose both research and engineering challenges. From a research standpoint, identifying an effective draft model remains an open problem, nearly as difficult as designing the target LLM itself. A draft model must closely approximate the target model’s token predictions while remaining much smaller for practical deployment. Toward this goal, a range of architectures has been explored: from lightweight MLPs or FFNs ~\parencite{stern2018blockwise,cai2024medusa,ankner2024hydra}, to RNN-based designs~\parencite{cheng2024apple_redrafter}, to more expressive Transformer-based decoding heads~\parencite{li2025eagle1,li2024eagle2}, which have demonstrated superior performance over simpler FFNs~\parencite{stern2018blockwise,cai2024medusa,gloeckle2024layerskip}.

More recently, self-attention-based autoregressive draft heads have gained traction due to their strong performance in future token prediction. Most of the latest high-performing SD systems~\parencite{li2024eagle2,ankner2024hydra,li2025eagle3,xiao2024clover2,zhang2025HASS} adopt a similar architecture: a self-attention layer that pools token embeddings and prior target states. Moreover, speculative decoding support remains limited in current LLM inference frameworks. At the time of writing, SGLang~\parencite{zheng2024SGlang} is the only publicly available framework that fully supports a leading SD method, i.e., EAGLE~\parencite{li2024eagle2,li2025eagle3}. Other frameworks~\parencite{huggingface2023assisted_generation,lmstudio} are either still under development or rely on disconnected (and often misaligned) draft and target models, resulting in marginal or even negative speedups. Much of the complexity of importing advanced SD models stems from non-standard architectural components and extensive customization for each new model integration.

Inspired by the success of the deep-shallow configuration in translation~\parencite{kasai2020deepshallow}, which reinterprets decoder-only LLMs as an encoder followed by a single cross-attention decoder, we explore whether a similarly minimal cross-attention structure can serve as a viable alternative to self-attention-based draft models. 

Unlike most existing self-attention-based SD solutions, we reduce a standard Transformer decoder~\parencite{vaswani2023transformers} to a minimal cross-attention structure without auxiliary layers.
It is shown to match the performance of state-of-the-art SD models for the same training data while maintaining architectural simplicity.
On the other hand, self-attention-based draft models generate queries and keys from the same hidden state, which complicates integration with autoregressive inputs and target outputs. To address this, existing work often require auxiliary pooling layers to handle heterogeneous features.
In contrast, our cross-attention architecture naturally handles different autoregressive states without pooling or custom fusion layers. Moreover, our cross-attention design enables efficient multi-token prediction during training, akin to “condensing”~\parencite{gao2021condenser,gao2021CoCondenser} future token information into the draft representation. To fully exploit this, we introduce a novel Two-Stage Block-Attention Training method that makes our architecture not only simple but effective.
Unlike prior Training-Time Test (TTT) based on self-attention~\parencite{li2025eagle3,zhang2025HASS}, our training scheme maintains constant memory usage and avoids the need to scale hidden states with the number of simulated steps, enabling full training of a 7B model on a single 24GiB GPU. We believe that our method offers a strong alternative to existing SD architectures, combining simplicity, familiarity, and practical efficiency.

\section{Related Work}
\noindent{}\textbf{Multi-token prediction training and SD:}
Initial speculative decoding work~\parencite{stern2018blockwise,sun2021early_gramar_err,xia2022firstSD} concentrated on tasks such as machine translation, where parallel mappings in word space make it easier to realize substantial speed gains.
These task-specific advances laid the foundation for more general-purpose SD methods~\parencite{cai2024medusa,ankner2024hydra,gloeckle2024layerskip}, which leverage parallel decoding heads for broader LLM applications.
A recent study~\parencite{lindsey2025biology} demonstrates that, in poetic text, hidden states—even from early positions—may already contain information about several upcoming tokens.
This insight aligns with empirical findings showing that LLM performance can be enhanced by multi-token prediction (MTP) \parencite{deepseekai2025deepseekv3} or by adopting more challenging objectives that inject additional contextual signals into model states~\parencite{gao2021condenser}.

While multi-token prediction has shown promise in prior work, state-of-the-art SD methods~\parencite{zhang2025HASS,li2025eagle3} continue to use autoregressive next-token prediction, aligning next-k tokens via step-by-step simulation during training, i.e., Training-Time Test, or TTT~\parencite{li2025eagle3}.
Furthermore, SD adaptation to every new LLM often entails training the SD model from scratch, amplifying the cost of training.
Although self-speculative methods~\parencite{Zhang2024draft_verify} address the co-training overhead by reusing the target model as the draft, they generally fall short in delivering good speed improvements~\parencite{zhong2024s3d}.

\noindent{}\textbf{Cross-attention heads for SD:} While it may seem natural to integrate draft and target model states through cross-attention, only limited prior work~\parencite{du2024glide,zimmer2024mixture,xiao2024clover2} have investigated this approach in the context of speculative decoding.
GLIDE with a CAPE~\parencite{du2024glide} employs a conventional cross-attention decoder that includes a self-attention sublayer.
In contrast, our approach eliminates half of the attention parameters by using a single-layer cross-attention module followed by an MLP.
Combined with an effective two-stage training scheme, this design advances cross-attention-based SD to achieve state-of-the-art speedups on the same training data scale~\parencite{li2024eagle2}, doubling the performance reported in~\parencite{du2024glide}.
MoA~\parencite{zimmer2024mixture}, which adds self-attention and mean aggregation layers on top of cross-attention to extract keys from the target model’s hidden states, further increases the complexity of the classic cross-attention module in the draft model.
As a result, MoA's speedup remains limited, and their evaluation of EAGLE-v2~\parencite{li2024eagle2} does not fully reflect the state-of-the-art speedup potential achievable at that data scale.
Clover-2~\parencite{xiao2024clover2} achieves effective speedups by incorporating cross-attention into one of several auxiliary layers.
Notably, its \textit{augment block} is solely used to improve first-token prediction.
In this work, we directly compare against Clover-2 and demonstrate that improved training efficiency alone allows us to surpass its speedups without introducing any inference-time overhead.

\section{Preliminaries}
Let $V$ be a discrete space over all possible tokens in the vocabulary, we model a target LLM of parameter $\Theta^*$ by the conditional distribution
$p_{n}(t) = \Pr(t_{n+1} = t \mid t_{1}, ..., t_{n}; \Theta^*)$ given context sequence $C = t_1, t_2, ..., t_{n}$ where $t_i \in V$ and the subscript $i$ denotes token positions.
In Transformer-based LLMs, the sampling of the next token \( t_{n+1} \sim p_n \) is conditioned on the preceding context \( C \) through the use of causal attention masks.
By modifying the attention masks, the dependency can be restricted to only a subset of tokens in $C$, enabling partial conditioning~\parencite{beltagy2020longformer,child2019sparseTransformers}.

In SD, the draft model \( q_{n}(t) = \Pr(\hat{t}_{n+1} = t \mid t_{1}, \ldots, t_{n}; \Theta) \) is optimized to approximate the target distribution, with \( \Theta \) optionally incorporating \( \Theta^* \) to enhance alignment.
During each SD iteration, a sequence of \( \gamma \) draft tokens \( \hat{t}_{n+1}, \hat{t}_{n+2}, \ldots, \hat{t}_{n + \gamma} \) is \textit{proposed}, and in lossless SD, only a contiguous prefix can be accepted.
In the \textit{verify} step, each proposed token \( \hat{t}_{n+i} \) is accepted with probability \( \min(1, p_{n+i-1}(\hat{t}_{n+i}) / q_{n+i-1}(\hat{t}_{n+i})) \) for \( i = 1, \ldots, \gamma \).  
At the first rejection position \( j \), or when \( j = \gamma + 1 \) without encountering any rejections, one additional token is sampled from normalized \( \max(0, p_{n+j-1} - q_{n+j-1}) \).
As a result, each SD iteration produces at least one new token, and here we denote the total number of accepted tokens $\tau \ge 1$.  
The above method ensures that accepted tokens are equivalent to those sampled from the target distribution~\parencite{leviathan2023sampling}.  
In the case of greedy decoding, this strategy effectively matches the top-1 tokens from the target and draft distributions, ensuring that the generated tokens exactly replicate the target model outputs.

The speedup potential comes from the fact that Transformers can verify multiple tokens in parallel in one forward pass with a time cost $T_v$ (often assumed to be constant within a small window).
Because the speed of SD-assisted generation is reflected by $\mathbb{E}[\tau] $ divided by the average iteration time cost $T = T_d + T_v$ where $T_d$ is the cost for drafting tokens, the per-iteration speedup, or \textit{improvement factor}~\parencite{leviathan2023sampling}, will be  
$\mathbb{E}[\tau] / (T_d/T_v + 1)$, which is seen as a proxy for the overall speedups.
Therefore, a high speedup requires both better acceptance lengths (when draft and target distributions align well) and low draft cost.

To reduce $T_d$, parallel multi-token SD methods proposes draft tokens in parallel~\parencite{cai2024medusa,gloeckle2024layerskip,zhong2024s3d,lin2024bita,xiao2024parallelspec,monea2023parallelspeculativesampling}.
However, the acceptance lengths of these models are generally worse than autoregressive SD methods~\parencite{li2025eagle1,li2024eagle2,zhang2025HASS}, although in the latter case $T_d$ is linearly proportional to $\gamma$.
Because many autoregressive models need only a single-layer draft model (compared to 32 layers when Llama 7B is the target model, essentially $T_d \ll T_v$) to be able to achieve $\mathbb{E}[\tau] > 3$ or even more, the speedups are generally more sensitive to the accuracy of predictions rather than iteration overheads.
To this end, most state-of-the-art draft models are autoregressive and are trained with the highest effective precisions (i.e., TF32)~\parencite{li2025eagle1,li2024eagle2,zhang2025HASS}.
To further maximize alignment between the draft and target models, these systems are also uniformly trained from scratch.
Together, these factors underscore the importance of addressing the overall training cost of speculative decoding.

\section{Methodology}
\begin{figure}
    \hspace{-0.25in}
    \includegraphics[width=1.1\linewidth]{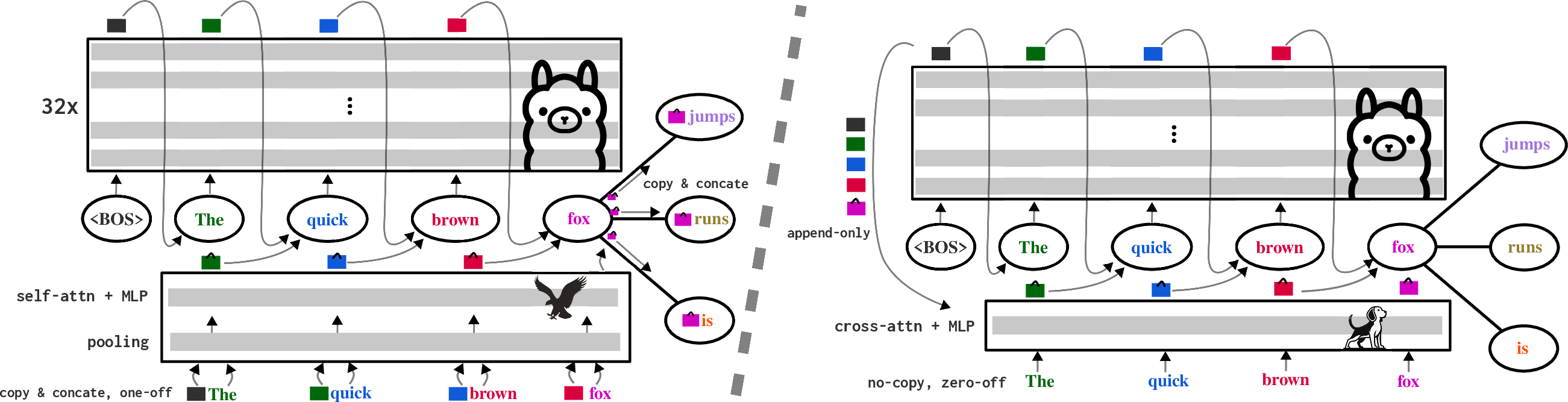}
    \vspace{0.1in}
    \caption{Comparison between EAGLE~\parencite{li2025eagle1,li2024eagle2} \textbf{(Left)} and our Beagle architecture \textbf{(Right)}. Square boxes denote higher-level states; a hat on top indicates states predicted by the draft model. Embedding layers are omitted for clarity, and colored words represent tokens generated at different positions. The right-side trees represent branched prediction via tree attention~\parencite{Miao_specInfer,cai2024medusa}.
    Using self attention,
    EAGLE requires auxiliary pooling layers and explicit copying of higher-level states for concatenation. 
    In contrast, Beagle adopts a standard training pipeline without offsets and avoids copying, simplifying draft modeling.
    }
    \label{fig:illustration}
\end{figure}

In this work, we propose an SD method, Budget EAGLE (Beagle), which does more accurate autoregressive predictions at inference time but utilizes multi-token parallel predictions during training to improve training efficiency and to condense more future information into draft model states.
Figure~\ref{fig:illustration} illustrates our model architecture at a high level and highlights its differences to a representative autoregressive SD method, EAGLE~\parencite{li2025eagle1,li2024eagle2}.

\subsection{Cross-Attention Draft Modeling}
Our draft model architecture, more specifically, is composed of a single-layer cross-attention Transformer block that maps lower-level context $t_1, ..., t_{n}$ to higher-level state $\mathbf{h}_{n}$ (for clarity, we omit layer details such as normalization, GQA, or positional embeddings):
\begin{equation}
\label{eq:TransformerBlock}
\begin{split}
  \mathbf{h}_n &= \operatorname{MLP}(\mathbf{y}_n) + \mathbf{y}_{n}\\
  \mathbf{y}_n &= \operatorname{CrossAttn}( \mathbf{h}_{1:n-1}, \mathbf{e}(t_n)) + \mathbf{e}(t_n)\\
\end{split}
\end{equation}
where $\mathbf{h}_i$ at any context position $i=1,...,n-1$ is expected to allow either target model top hidden states (we name them \textit{true states}) or the autoregressively generated states from the draft model itself.
Furthermore, $\mathbf{e}: t \rightarrow \mathbb{R}^d$ is the embedding layer, and $\operatorname{MLP}$ is a point-wise feed-forward layer.

For the cross-attention sublayer, more specifically, the query, key, and values are processed as follows:
\begin{equation}
\label{eq:cross_attn}
\begin{split}
  \operatorname{Q}^{(h)}_i\;, \operatorname{K}^{(h)}_j\;, \operatorname{V}^{(h)}_j &= W^T_{h,Q} \mathbf{e}(t_i) \;, W^T_{h,K} \mathbf{h}_j \;, W^T_{h,V} \mathbf{h}_j \\
  s^{(h)}_{i, j} &= \operatorname{Softmax}_j(\langle \operatorname{Q}^{(h)}_i, \operatorname{K}^{(h)}_j \rangle / \sqrt{d_h}) \\
  \mathbf{o}_i^{(h)} &= \sum_j \operatorname{Mask}^{(h)}_{i,j} \cdot s^{(h)}_{i, j} \cdot V_j^{(h)} \\
  \mathbf{y}_i   &=  W^T_O [\mathbf{o}^{(1)}_i; \mathbf{o}^{(2)}_i; ..., \mathbf{o}^{(H)}_i]_{d\times 1}
\end{split}
\end{equation}
where weights $W_{h, Q}, W_{h, K}, W_{h, V} \in \mathbb{R}^{d \times d_h}$ and $W^T_O \in \mathbb{R}^{d \times d}$ where
$d$ and $d_h$ are model and head hidden dimensions, respectively.
Unlike causal self attention, the cross-attention mask here has to ensure ``diagonal scores'' are also masked, i.e., $\operatorname{Mask}_{i,j} = 0$ for $j \ge i$ given query at position $i$.
Using constant-space cross-attention masks, we can allow predicting multiple future tokens during training (see Section~\ref{sect:2stage_training}).

Finally, our draft model can be defined by $q_{n}(t) = \operatorname{Softmax}(\mathbf{z}_n)$ where logits $\mathbf{z}_n = \mathbf{e}^{-1}(\mathbf{h}_n)$ and the language model head is a linear mapping $\mathbf{e}^{-1}: \mathbb{R}^d \rightarrow \mathbb{R}^V$.

As seen in Eq.~\ref{eq:TransformerBlock}, we replace the commonly used self-attention layer with a single cross-attention layer.
However, this cross-attention Transformer block is used in a \textit{non-standard} causal fashion to decode draft tokens autoregressively during inference:
At query position $i$, the computation of $s_{i,j}$ and $\mathbf{o}_i$ in Eq.~\ref{eq:cross_attn} can reuse existing $K_j, V_j$ if they are cached for all $j < i$,
and we only need to append new states $K_i, V_i$ to KV-cache when we query state $\mathbf{h}_{i+1}$.
Lastly, we need to reset KV-cache to only contain true states at the end of SD iteration.

These changes eliminate the need to use any auxiliary pooling layers because we can handle low- and high-level states via different queries and key/value space.
As a result, it also avoids copying and concatenating high-level states to feed the draft model as next inputs (see Figure~\ref{fig:illustration}), leading to greater memory locality.
We will show, with effective training, this simplified architecture can perform evenly or better than more complex architectures commonly seen in recently developed SD models~\parencite{li2024eagle2, xiao2024clover2}.

\subsection{Two-Stage Block-Attention Training}
\label{sect:2stage_training}
Many autoregressive speculative decoding methods are trained to predict only the immediate next token following the training-data token, which is effectively equivalent to training a draft LLM using the Next Token Prediction (NTP) objective, conditioned on the target model’s runtime hidden states.
However, one-step NTP does not explicitly capture the actual inference dynamics in speculative decoding, particularly when the draft model starts to rely on its own predicted hidden states during autoregressive inference.
Within a draft window, prediction errors and accumulated noise can cause the draft model’s behavior to diverge from the target distribution used during training, leading to a suboptimal speed.

As a result, many recent speculative training methods have adopted the Training-Time Testing (TTT) scheme to explicitly allow potentially inaccurate predictions and to train on the simulated inference data, effectively ``unrolling'' for multiple steps during training.
However, this is at the expense of much longer training time, which we believe would not be suitable for the entire training cycle.

Instead, during the \textbf{early stage}, we propose to predict multiple future tokens $\hat{t}_{n+1}, \hat{t}_{n+2}, ..., \hat{t}_{n+k}$ and feed them to Transformer in parallel. And only in the \textbf{late stage}, we apply training-time simulation.
Compared to self-attention heads, we will show this leads to reduced training overheads as well.

\begin{figure}
    \centering
    \includegraphics[width=1\linewidth]{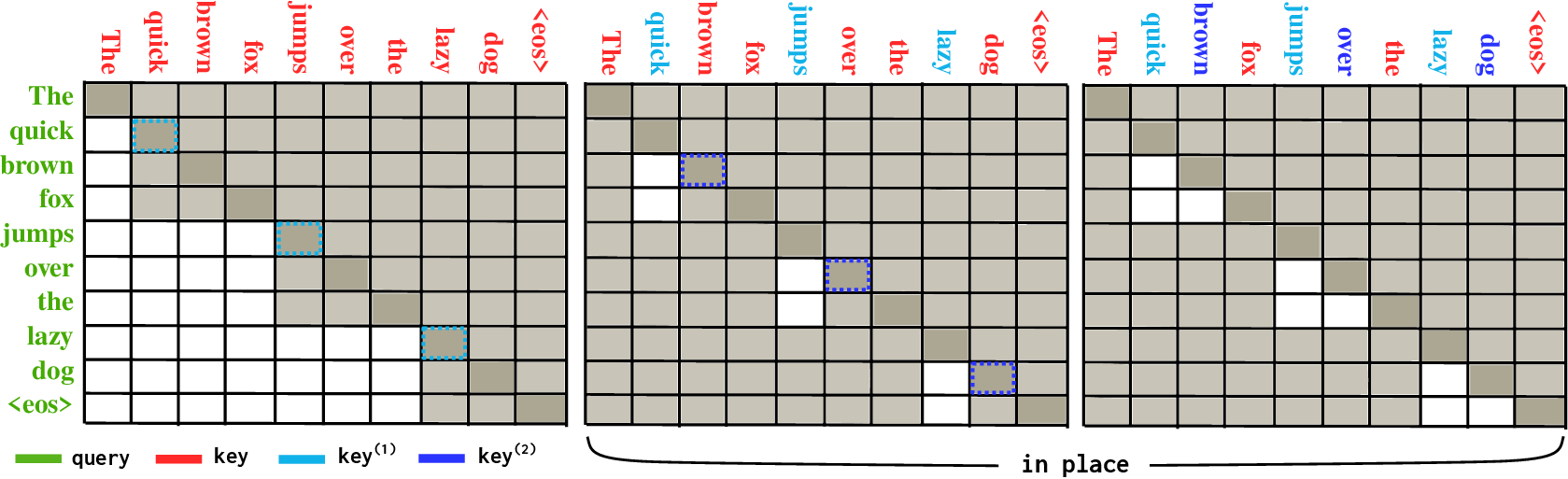}
    \caption{The cross-attention masks used during training for draft model heads. \textbf{Left one (early stage block attention):} Query states are derived directly from token embeddings, and keys are from high-level states. An \textit{inverse} block attention starting at a random offset with a fixed window hides local keys from a query, encouraging the model to condense more information on future tokens. \textbf{Right two (after simulation step 1 and step 2, late stage):} In the late-stage training, we unroll newly predicted states to accurately simulate inference during training. Unlike Training-Time Test with self attentions, this method requires no new queries to be generated, and only needs one-step attention memory allocation for in-place adding of next-predicted keys.}
    \label{fig:block_TTT}
\end{figure}

Denote the model prediction distribution to a $i$-step ahead future token $t_{n+i}$ as
\begin{equation}
\label{eq:future_dist}
  q^{(i)}_{n}(t) = \Pr(\hat{t}_{n+i} = t \mid t_{1}, ..., t_{n}; \Theta). 
\end{equation}
where $q^{(1)}_{n}(t) = q_{n}(t)$.
We mask out continuous windows of tokens in the attention mask (starting at a random minor offset).
Specifically, at a window of size $k$ starting at $n$, and for a query at position $n+i, i = 1,...,k$, the corresponding local future keys at $n + j, 1 \le j \le k$ are all masked out.
This masking results in a block attention matrix as shown in Figure~\ref{fig:block_TTT} (the left-most attention mask), different from usual block attentions where only local tokens are seen~\parencite{jiang2023mistral7b}, we may dub it \textit{inverse} block attentions because local tokens are masked out.

Other than encouraging the draft model to contain representations for multiple future tokens, we also make sure each query state will be utilized to backpropagate losses, thus maximizing sampling efficiency in block attentions. As such, we define our early-stage loss as
\begin{equation}
\label{eq:early_loss}
    \mathcal{L}_{early} = - \frac 1N \sum_{n \in w_0} \sum_{j=1}^k \mathbb{E}_{t \sim p_{n+j}} [\log q^{(j)}_n(t)]
\end{equation}
where $k$ is the window size and $j$ represents the query position relative to window start positions $w_0 = \{\epsilon + w \cdot k: w=1,..., N\}$ with a random offset $\epsilon \in [0, k)$. 
Additionally, the maximum number of windows $N$ is selected to cover all training inputs for maximum sample efficiency.

For late-stage training, we start to simulate multiple steps of inference (illustrated in the right two attentions in Figure~\ref{fig:block_TTT}).
Specifically, the predicted states at the start of the window ($i = 1$) are saved and expanded to the next step, where the initial attention mask is also extended and allows next queries in the same window to access updated states.
And after the second step, the newly predicted states are again saved, but via in-place modifications with a correspondingly modified attention mask.
This process is repeated in a single training iteration until a total simulation steps $s$.
Unlike the linear expansions seen in self-attention training~\parencite{zhang2025HASS,li2025eagle3}, here our inference simulation consumes constant space and does not need to unroll queries during training.

At simulation step $i$, denote the $j$-step ahead future draft distribution as
\begin{equation}
\label{eq:future_dist_unroll}
  q^{(i, j)}_{n}(t) = \Pr(\hat{t}_{n+j} = t \mid t_1, ..., t_n, \hat{t}_{n+1}, ..., \hat{t}_{n+i-1}; \Theta)
\end{equation}
where we have predicted tokens up to $n + i - 1$, and are predicting token at $n+j$ where $j \ge i$.
The true context of states from training data is still fixed back at position $n$.
The general form of loss considering simulation is, for $1 \le s \le l \le k$,
\begin{equation}
\label{eq:late_loss}
    \mathcal{L}(s, l, \beta) = - \frac 1N \sum_{n \in w_0} \sum_{i=1}^s \sum_{j=i}^{l} \beta_{i,j} \cdot \mathbb{E}_{t \sim p_{n+j}} [\log q^{(i, j)}_n(t)].
\end{equation}
Here, \( s \) denotes the maximum number of simulated steps, and \( l \) indicates the maximum lookahead position. The associated weight \( \beta_{i,j} > 0 \) is used to re-weight loss terms.

We select \( s = k \), \( l = i \), and \( \beta^*_{i,j} = k - i + 1 \) to define our late-stage loss as \( \mathcal{L}_{\text{late}} = \mathcal{L}(k, i, \beta^*) \).  
We provide justifications in Appendix~\ref{sect:justification} showing that this formulation serves as a surrogate loss that approximates the acceptance length during SD inference, and offers a better approximation than the early-stage loss toward the end of training.

\section{Experiments}
\label{sect:experiments}

\subsection{Experimental Setups}
\label{sect:experimental_setups}
In this work, we limit our baselines to lossless decoding methods and focus on single-batch greedy decoding. The optimization for throughput and speculative sampling is left for future work.

\noindent{}\textbf{Baselines:}
We consider popular target models in this domain: Vicuna (7B), LLaMA-2 (7B), and LLaMA-3 (7B).
We use HuggingFace TGI (Text Generation Inference) for baseline SD, paired with JackFram LLaMA-68M and Vicuna-68M as used in other work~\parencite{Miao_specInfer,yang2024mcsd}.
We consider a popular inference-time parallel decoding SD method, Medusa~\parencite{cai2024medusa}, with official Vicuna weights.
Moreover, a representative zero-memory-overhead self-speculative method~\parencite{Zhang2024draft_verify} with official LLaMA-2 weights is also added.
Importantly, we include Clover 2~\parencite{xiao2024clover2}, which is reportedly the most efficient model based on cross attention.
EAGLE v1 and v2 series represent the best open models using the same training data scales. EAGLE v2, which adopts a dynamic draft attention tree, constantly performs better~\parencite{li2024eagle2}, so we always include EAGLE-v2 and adopt the same dynamic tree method.
We do not include EAGLE-v3 (which is trained on 8× more data) or much more expensive full-stage TTT training approaches~\parencite{zhang2025HASS}, as our focus is on exploring efficient training strategies and architectural alternatives under comparable data scales.

\noindent{}\textbf{Datasets:}
We limit our training dataset to only ShareGPT~\parencite{sharegpt}, which is composed of over 60K conversational dialogues from ChatGPT and its users.
This is to align with other baselines~\parencite{cai2024medusa,li2025eagle1,li2024eagle2,xiao2024clover2} with the same amount of training data.
Our inference datasets cover multi-turn general conversational benchmark MT-Bench~\parencite{zheng2023mt_bench}, reasoning task GSM-8K~\parencite{cobbe2021gsm1k}, and summarization task CNN-Daily~\parencite{hermann2015cnn_daily} with a subset of 1,000 samples following \cite{Zhang2024draft_verify}.
In total, a full evaluation run for one system covers more than 2,100 inputs (including conversation turns), and our measurement values (other than peak memory) are aggregated averages.

\noindent{}\textbf{Inference and Training:}
All implementations are based on HuggingFace Transformers~\parencite{wolf2020huggingfaces} contained in the same Docker environment\footnote{We use the official \texttt{pytorch:2.6.0-cuda11.8-cudnn9-runtime} as our base Docker image.} and are running in PyTorch eager mode with BF16 precisions during inference.
Our evaluation framework also makes sure each system is running inference against the same data, but we follow the training prompt formats of each baseline for maximum speed.
The inference jobs are run on two grades of servers using single-threaded executions: an NVIDIA A6000 Ada node and an Amazon AWS instance with A10G GPUs.

Similar to trained baseline models, our model is trained with mixed precisions where model weights are TF32 while target model states are preserved in half precisions, which are generated offline to minimize training time and GPU memory usage.
If not specified otherwise, we train 20 epochs maximum, where the first 10 epochs we use the early-stage training strategy and the rest use the late-stage training strategy.
The detailed training configurations of our different settings can be found in Appendix~\ref{sect:training}.
As official EAGLE weights are trained from an unknown number of epochs, we replicate EAGLE-v2 with the same number of training epochs to ensure comparable results. Also, we align the EAGLE dynamic attention tree with the same hyperparameters (i.e., depth=1+6, topk=10, and candidates to be accepted per draft iteration is set to 60).

\subsection{Results}

\begin{table}[]
\centering
\caption{Speed and memory comparisons among different models (A6000 Ada). The left two columns specify Target and Draft models, where V, L2, and L3 represent Vicuna, LLaMA-2, and LLaMA-3, respectively. The metrics Spu, $\tau$, and M represent Speedup, acceptance length, and GPU peak Memory usage.}
\label{tab:main}
\resizebox{\columnwidth}{!}{%
\begin{tabular}{ll|rrrr|rrrr|rrrr}
\toprule
\multirow{2}{*}{T} & \multirow{2}{*}{D} & \multicolumn{4}{c}{\bf MT-Bench} & \multicolumn{4}{c}{\bf GSM-8K} & \multicolumn{4}{c}{\bf CNN-Daily} \\
 &  & \multicolumn{1}{l}{Speed} & \multicolumn{1}{l}{Spu} & \multicolumn{1}{l}{$\tau$} & \multicolumn{1}{l}{M$\downarrow$} & \multicolumn{1}{l}{Speed} & \multicolumn{1}{l}{Spu} & \multicolumn{1}{l}{$\tau$} & \multicolumn{1}{l}{M$\downarrow$} & \multicolumn{1}{l}{Speed} & \multicolumn{1}{l}{Spu} & \multicolumn{1}{l}{$\tau$} & \multicolumn{1}{l}{M$\downarrow$} \\
\midrule
\multirow{6}{*}{V} & None & 35.2 & 1.0 & 1.0 & 13.7 & 35.3 & 1.0 & 1.0 & 13.5 & 34.2 & 1.0 & 1.0 & 14.1 \\
 & TGI & 56.8 & 1.6 & 3.9 & 13.1 & 60.1 & 1.7 & 2.9 & 12.9 & 56.4 & 1.6 & 2.9 & 13.6 \\
 & Medusa & 71.2 & 2.0 & 2.1 & 15.0 & 83.5 & 2.4 & 2.6 & 14.9 & 63.0 & 1.8 & 2.0 & 15.3 \\
 & Clover 2 & 53.3 & 1.5 & \bf 4.1 & 16.0 & 57.2 & 1.6 & 4.2 & 16.0 & 47.2 & 1.4 & 3.6 & 16.4 \\
 & EAGLE v2 & 103.4 & 2.9 & 4.0 & 14.5 & \bf 118.4 & \textbf{3.4} & \bf 4.6 & 14.3 & \bf 87.6 & \textbf{2.6} & \bf 3.6 & 14.8 \\
 & Beagle (Ours) & \bf 104.6 & \textbf{3.0} & \bf 4.1 & 13.5 & 108.5 & 3.1 & 4.3 & 13.3 & 82.0 & 2.4 & 3.4 & 14.0 \\
\midrule
\multirow{6}{*}{L2} & None & 34.9 & 1.0 & 1.0 & 13.8 & 35.3 & 1.0 & 1.0 & 13.3 & 34.8 & 1.0 & 1.0 & 14.2 \\
 & Self-Spec & 34.8 & 1.0 & 1.7 & 13.7 & 34.1 & 1.0 & 1.7 & 13.1 & 34.6 & 1.0 & 1.9 & 14.2 \\
 & TGI & 46.0 & 1.3 & 3.3 & 13.1 & 47.6 & 1.3 & 2.6 & 12.9 & 40.8 & 1.2 & 2.6 & 13.6 \\
 & EAGLE v2 & 104.9 & \textbf{3.0} & 4.0 & 15.5         & \bf 111.6 & \textbf{3.2} & 4.4 & 15.4        & \bf 89.8 & \textbf{2.6} & \bf 3.7 & 15.9 \\
 & Beagle (Ours) & \bf 106.2 & \textbf{3.0} & \bf 4.1 & 13.5    & 111.4 & \textbf{3.2} & \bf 4.5 & 13.2                 & 85.4 & 2.5 & 3.6 & 14.0 \\
\midrule
\multirow{3}{*}{L3} & None & 32.3 & 1.0 & 1.0 & 15.6 & 32.5 & 1.0 & 1.0 & 15.5 & 33.3 & 1.0 & 1.0 & 15.8 \\
 & EAGLE v2 & \bf 80.2 & \textbf{2.5} &\bf 3.6 & 17.8         & 83.0 & \textbf{2.6} & 3.9 & 17.7        & \bf 69.6 & \textbf{2.1} & \bf 3.4 & 18.1 \\
 & Beagle (Ours) & 79.2 & \textbf{2.5} & 3.5 & 15.7    & \bf 83.5 & \textbf{2.6} &\bf 4.0 & 15.5                 & 67.2 & 2.0 & 3.2 & 16.0 \\
\bottomrule
\end{tabular}
}%
\end{table}

Table~\ref{tab:main} (and \ref{tab:main2} in the Appendix) compares both speeds (in tokens per second) and peak memory usage among systems across two different grades of GPUs.
Our system, Beagle, has shown a similar efficiency level to EAGLE v2, where both are trained with the same training data for 20 epochs.
However, our memory overhead on top of the target model is minimal, while EAGLE consumes 10\% to 15\% more GPU memory.

On the other hand, Self-Spec using self-speculative decoding consumes no additional memory and does generate greater-than-one acceptance lengths (thus better than no-SD), but it leads to no speed improvements because its draft model consists of multiple layers of overheads.
For the same reason, Clover 2 -- with various augment layers which help to achieve much better acceptance rates -- obtains only around 1.5x speedups.
Due to the lack of co-training and model alignment, baseline SD (TGI) adds little speedup as well. 
Finally, Medusa, using parallel decoding at inference, has suboptimal acceptance lengths compared to autoregressive models such as EAGLE v2 and ours.

\begin{figure}
    \centering
    \includegraphics[width=1\linewidth]{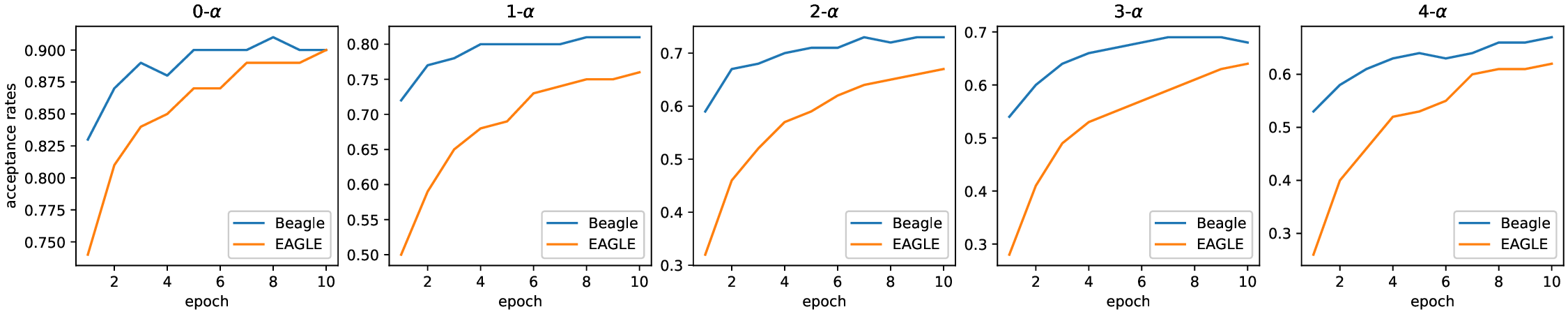}
    \caption{\textbf{Early-stage} acceptance rates at different draft steps (step-$\alpha$) during the first 10 epoch training process (evaluated on MT-Bench). Our model (Beagle) uses the early-stage loss based on multi-token predictions. At this stage, our training efficiency is consistently better than EAGLE (v1/v2)~\parencite{li2025eagle1,li2024eagle2}}
    \label{fig:training_efficiency}
\end{figure}
\begin{figure}
    \centering
    \includegraphics[width=0.7\linewidth]{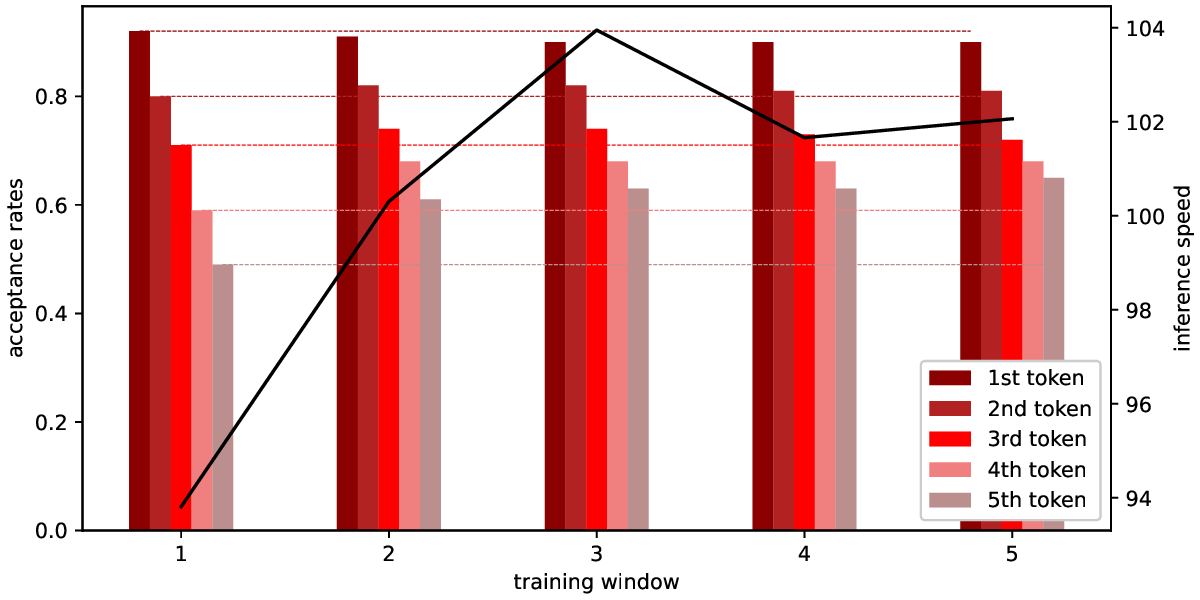}
    \caption{\textbf{Early-stage} token acceptance rates at different positions and corresponding inference speeds (evaluated on MT-Bench). We vary the window length from 1 to 5 for five early-stage training settings.
    Multi-token prediction (using $\mathcal{L}_{early}$) with a proper window width (optimal width achieved at 3) improves further-step token acceptance rates, generally enhancing inference speeds.
    }
    \label{fig:compare_winds}
\end{figure}

Other than time and memory efficiency, our having fewer parameters than EAGLE also enables much greater training efficiency in the early stage.
As shown in Figure~\ref{fig:training_efficiency}, although we converge to a similar acceptance rate for the first token, our system shows consistently better performance during the early stage of training.
Additionally, according to Figure~\ref{fig:compare_winds}, the multi-token prediction loss (Eq.~\ref{eq:early_loss}) used in the early stage leads to better training results and improved future token predictions.
Moreover, it consumes no more data than EAGLE, utilizing the Transformer's parallel forwards advantage.

\subsection{Justifications for Two-Stage Training}
\label{sect:main_justifications}
We conduct experiments to verify our interpretations for two-stage losses in Appendix~\ref{sect:justification}, i.e., (1) the early-stage loss is a worse surrogate (but trains more efficiently);
(2) and the late-stage loss corresponds to inference efficiency more precisely (although spending more compute on each training iteration).

As shown in Figure~\ref{fig:compare_winds}, agnostic to training window sizes, the future token acceptance rates constantly show further degradations over distances.
But unlike the strict assumption we have in Appendix~\ref{sect:justification}, the multi-token training loss $\mathcal{L}_{early}$ can actually ``bend'' the decline curves to form a slower slop as window size enlarges, which does not necessarily improve the 1st token acceptance rates but instead enhancing the overall acceptance among all future tokens -- leading to a better speed than single-token prediction training baseline (window=1) and also justifying our early-stage training loss.

However, there is an optimal window size (at 3) which can lead to the best end speed number.
This is likely a trade-off between focusing on early tokens or on late tokens -- although a large window helps preserve degradation on future token predictions, it does hurt the 1st token acceptance rates beyond a window size of 2, at the meantime, the 1st token acceptance rate is crucial (as it is weighted the most in the expected acceptance length as shown in Appendix~\ref{sect:justification}).

\begin{figure}
    \centering
    \includegraphics[width=0.95\linewidth]{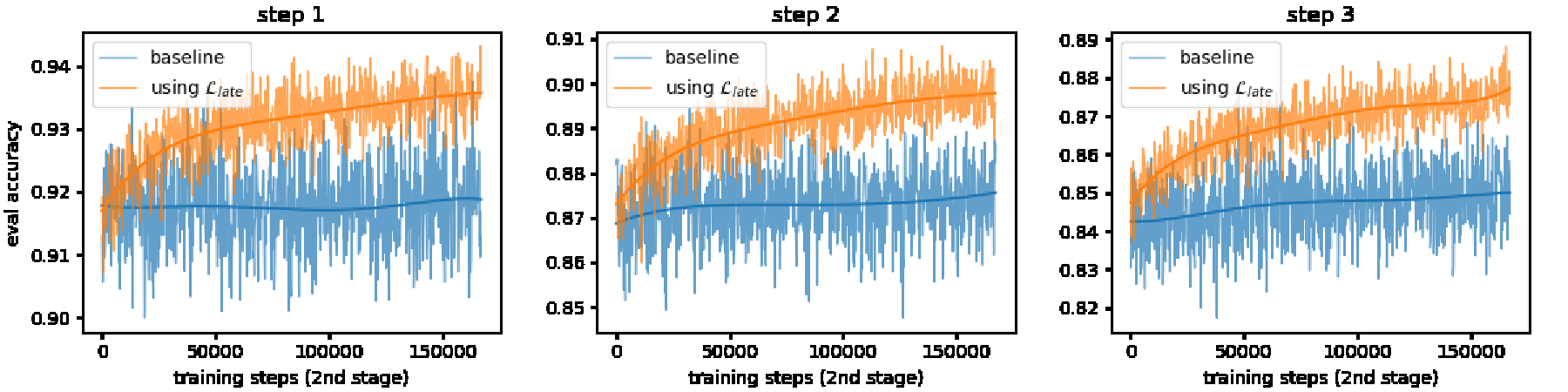}
    \caption{
    The \textbf{late-stage} (10th- to 20th-epoch) draft model prediction accuracy changes using different training losses (the validation set during training is a partial MT-Bench data). The orange lines correspond to the model trained with our proposed late-stage loss $\mathcal{L}_{late}$, and the blue baselines are when the model continues to be trained with early-stage loss $\mathcal{L}_{early}$. Due to the high variance of accuracy changes during late-stage training, we also highlight the interpolated smooth curves.
    }
    \label{fig:2nd_stage}
\end{figure}

Figure~\ref{fig:2nd_stage} justifies the necessity of our late-stage loss.
Although our late-stage training loss requires simulating each query token for multiple steps to train on policy (thus obviously adding linear time overheads with respect to steps), it is necessary to obtain better prediction capabilities after training.
Towards the end of the training, the first-stage loss offers minimal accuracy improvements over time (blue lines in Figure~\ref{fig:2nd_stage}), while the late-stage training loss can keep advancing prediction accuracies notably for all shown unrolled steps.
As a result, we consider our late-stage training to be a complementary and necessary addition to the early-stage training.

Finally, at the end of training, our late-stage surrogate loss \( \mathcal{L}_{late} \) is shown (in Appendix~\ref{sect:justification}) to have an almost constant bound w.r.t. the approximated acceptance length.  
In contrast, many existing SD training approaches apply uniform weighting to predicted tokens across different steps. 

\section{Conclusion}
In this work, we present a novel cross-attention-based speculative decoding (SD) modeling along with an effective, well-grounded two-stage training scheme built on block attention mechanisms.  
Our method employs a simpler and less tailored architecture without auxiliary layers, having an effectively improved early-stage training efficiency and constant GPU memory usage during simulated inference.  
With improved training strategies, we demonstrate -- for the first time -- that cross-attention models can match the performance of state-of-the-art EAGLE-v2 self-attention architecture on the same training data.
We believe this work opens new research directions for exploring more diverse architectures and applications in SD, e.g., optimizing vision-language models (VLMs) for vision tasks.

\section{Limitations}
\label{sect:limitations}
Our hypothesis that the $i$-th ahead token follows a geometric degradation in accuracies \textit{if} these tokens are predicted in parallel (Eq. ~\ref{eq:assumption}) may not strictly reflect real observations.
However, this does not undermine our major conclusions and the necessity of our proposed two-stage training because we have provided sufficient arguments and empirical results in Section~\ref{sect:main_justifications}.
%
Secondly, our work is limited to exploring efficient training strategies and architectural alternatives under comparable data scales. As a result, systems achieving greater speedups by training with a different scale of data (e.g., EAGLE-v3 using 8x more data) or with more expenses are not compared in this work.
Finally, we have conducted training and experiments only on smaller-scale models because we are limited by resources to train larger LLMs with permissive hardware, and the selection of target models is also largely restricted by commonly used model checkpoints shared among our evaluated baselines.
We believe scaling our effectiveness to different model sizes is an orthogonal topic and can be left to future work.

\begin{ack}
This collaborative work was fully funded by LG Electronics, Toronto AI Lab. We sincerely thank everyone for their patience and support throughout this research. Special thanks go to Paria Nejat, Chul Lee, and Kevin Ferreira for their generous assistance with hardware and resource allocation. We also greatly appreciate Manasa Bharadwaj, one of the authors, for her coordination and resource management during the final stage of the project.
\end{ack}

\printbibliography

\appendix
\clearpage
\section{Appendix}

\subsection{Supplementary Evaluation Table (A10G)}
\begin{table}[h]
\centering
\caption{Speed and memory comparisons among different models (A10G). The left two columns specify Target and Draft models, where V, L2, and L3 represent Vicuna, LLaMA-2, and LLaMA-3, respectively. The metrics Spu, $\tau$, and M represent Speedup, acceptance length, and GPU peak Memory usage.}
\label{tab:main2}
\resizebox{\columnwidth}{!}{%
\begin{tabular}{ll|rrrr|rrrr|rrrr}
\toprule
\multirow{2}{*}{T} & \multirow{2}{*}{D} & \multicolumn{4}{c}{\bf MT-Bench} & \multicolumn{4}{c}{\bf GSM-8K} & \multicolumn{4}{c}{\bf CNN-Daily} \\
 &  & \multicolumn{1}{l}{Speed} & \multicolumn{1}{l}{Spu} & \multicolumn{1}{l}{$\tau$} & \multicolumn{1}{l}{M$\downarrow$} & \multicolumn{1}{l}{Speed} & \multicolumn{1}{l}{Spu} & \multicolumn{1}{l}{$\tau$} & \multicolumn{1}{l}{M$\downarrow$} & \multicolumn{1}{l}{Speed} & \multicolumn{1}{l}{Spu} & \multicolumn{1}{l}{$\tau$} & \multicolumn{1}{l}{M$\downarrow$} \\
\midrule
\multirow{6}{*}{V} & None & 13.3 & 1.0 & 1.0 & 13.7 & 13.1 & 1.0 & 1.0 & 13.5 & 13.0 & 1.0 & 1.0 & 14.1 \\
 & TGI & 23.6 & 1.8 & 4.0 & 13.1 & 23.8 & 1.8 & 2.9 & 12.9 & 21.7 & 1.7 & 2.7 & 13.6 \\
 & Medusa & 27.2 & 2.0 & 2.1 & 15.0 & 32.3 & 2.5 & 2.6 & 14.9 & 24.6 & 1.9 & 2.0 & 15.3 \\
 & Clover 2 & 21.6 & 1.6 & 4.0 & 16.0 & 21.8 & 1.7 & 4.2 & 16.0 & 16.8 & 1.3 & 3.5 & 16.4 \\
 & EAGLE v2 & \bf 40.0 & \bf 3.0 & 3.9 & 14.5               & \bf 46.4 & \bf 3.5 &\bf 4.7 & 14.3 & 30.7 & \bf 2.4 &\bf 3.6 & 14.8 \\
 & Beagle (ours) & 39.8 & \bf 3.0 &\bf 4.1 & 13.5          & 41.9 & 3.2 & 4.3 & 13.3 & \bf 31.6 & \bf 2.4 & 3.4 & 14.0 \\
\midrule
L2 & None & 13.2 & 1.0 & 1.0 & 13.8 & 13.3 & 1.0 & 1.0 & 13.3 & 13.0 & 1.0 & 1.0 & 14.2 \\
 & Self-Spec & 13.8 & 1.0 & 1.7 & 13.7 & 13.2 & 1.0 & 1.7 & 13.1 & 13.7 & 1.0 & 1.9 & 14.2 \\
 & TGI & 17.9 & 1.4 & 3.2 & 13.1 & 19.0 & 1.4 & 2.6 & 12.9 & 15.8 & 1.2 & 2.8 & 13.6 \\
 & EAGLE v2 & 41.7 & \bf 3.2 &\bf 4.1 & 15.5 & 42.8 & 3.2 & \bf 4.4 & 15.4 &\bf 34.7 & \bf 2.7 &\bf 3.7 & 15.9 \\
 & Beagle (ours) &\bf 41.9 & \textbf{3.2} &\bf 4.1 & 13.5 & \bf 43.5 & \bf 3.3 & \bf 4.4 & 13.2 & 32.7 & 2.5 & 3.6 & 14.0 \\
\midrule
L3 & None & 12.4 & 1.0 & 1.0 & 15.6 & 12.7 & 1.0 & 1.0 & 15.5 & 12.5 & 1.0 & 1.0 & 15.8 \\
 & EAGLE v2 &\bf 33.7 & \bf 2.7 &\bf 3.6 & 17.8 & \bf 34.9 &\bf 2.7 & 3.9 & 17.7 &\bf 28.9 &\bf 2.3 &\bf 3.4 & 18.1 \\
 & Beagle (ours) &33.2 & \bf 2.7 &\bf 3.6 & 15.7 & \bf 34.9 & \bf 2.7 & \bf 4.0 & 15.5 & 27.9 & 2.2 & 3.2 & 16.0 \\
\bottomrule
\end{tabular}
}%
\end{table}

\subsection{Interpretations of Two Stage Losses}
\label{sect:justification}
For better SD inference performance, we are essentially optimizing the expected acceptance length $L$ within a window $k$, that is,
\begin{equation}
\begin{split}
\mathbb{E}[L] &= \sum_{\ell=1}^{k} \Pr(L\ge \ell) \qquad \text{(tail-sum formula)} \\
&= \sum_{i=1}^{k} \exp(\sum_{j=1}^{i} \log \alpha^{(j)})
\end{split}
\end{equation}
where $\alpha^{(i)}$ denotes the expected acceptance rate at position $i$.
Although this objective can be directly modeled as a loss function using negative $\log\mathbb{E}[L]$ and calculating $\operatorname{logsumexp}$ of accumulated $\log\alpha^{(i)}$ values for each simulated step, the numerical issue arise due to the large differences of magnitudes of values in different steps.
However, as long as the training is effective (it is a reasonable assumption because cross-entropy loss terms are also maximizing the data log likelihoods), we may assume $\log \alpha^{(i)} \rightarrow 0$ towards the end of training.
Then, by first-order Taylor expansion,
\begin{equation}
\label{eq:approx_objective}
\begin{split}
\mathbb{E}[L] \approx J &= \sum_{i=1}^{k} (1 + \sum_{j=1}^{i} \log \alpha^{(j)}) \\
&= \sum^k_{i=1} (k - i + 1) \log \alpha^{(i)} + k.
\end{split}
\end{equation}
This time, the RHS objective $J$ in Eq.~\ref{eq:approx_objective} is a more numerically stable objective.

On the other hand, we hypothesize that the $i$-th ahead token follows a geometric degradation in accuracies \textit{if} these tokens are predicted in parallel:\footnote{We do not hypothesize a degradation in autoregressive predictions because \cite{li2025eagle3} show that the acceptance rates can be maintained very effectively given enough training data.}
\begin{equation}
\label{eq:assumption}
\alpha^{(i)} = r^{\, i - 1} \alpha^{(1)}
\end{equation}
where the degradation rate $r = r(n, \Theta) < 1$ is a variable depending on the draft model of parameters $\Theta$ given a context prior to position $n$.

With the draft distribution notation $q^{(i, j)}_n(t)$ in Eq.~\ref{eq:future_dist_unroll}, our general loss function in Eq.~\ref{eq:late_loss} can be rewritten as (for one window with non-branching prediction)
\begin{equation}
\label{eq:surrogate}
\begin{split}
\mathcal{L}(s, l, \beta) &= - \sum_{i=1}^s \sum_{j=i}^{l} \beta_{i,j} \cdot \mathbb{E}_{t \sim p_{n+j}} [\log q^{(i, j)}_n(t)] \\
&\ge - \sum_{i=1}^s \sum_{j=i}^{l} \beta_{i,j} \cdot \log \mathbb{E}_{t \sim p_{n+j}} [ q^{(i, j)}_n(t)] \qquad \text{(Jensen's inequality)}\\
&= - \sum_{i=1}^s \sum^{l}_{j=i} \beta_{i,j} \cdot \log (r^{(j - i)} \alpha^{(i)}) \\
&= - \sum_{i=1}^s \sum^{l}_{j=i} \beta_{i,j} \cdot [(j - i)\log r + \log \alpha^{(i)}] \\
&\ge - \sum_{i=1}^s \sum^{l}_{j=i} \beta_{i,j} \log \alpha^{(i)} \qquad (s,l \text{ are hyperparameters})\\
\end{split}
\end{equation}
where equality holds when $i = j$.

We may denote $q^{(j)}_n(t) = q^{(1, j)}_n(t)$ according to Eq.~\ref{eq:future_dist} and \ref{eq:future_dist_unroll}, the early loss is then
\begin{equation}
\label{eq:early_bound}
\mathcal{L}_{early} = \mathcal{L}(1, k, \mathbf{1}) \ge - k \log \alpha^{(i)} = -J + \sum^k_{i=2} (k - i + 1) \log \alpha^{(i)} + k \ge -J
\end{equation}
which can be seen as a surrogate to the objective in Eq.~\ref{eq:approx_objective}.

Alternatively, when we simulate the exact autoregressive decoding (predicting the immediate next token in many steps) by assigning $s = k, l = i$ and $\beta^*_{i,j} = k - i + 1$, it becomes our proposed late-stage loss $\mathcal{L}_{late}$ and it is also a surrogate to the objective because
\begin{equation}
\label{eq:late_bound}
\mathcal{L}_{late} = \mathcal{L}(k, i, \beta^*) \ge - \sum_{i=1}^k (k - i + 1) \log \alpha^{(i)} = -J + k.
\end{equation}
Compared to Eq.~\ref{eq:late_bound} where there is an almost constant gap (up to a \textit{Jensen's Gap} due to Eqs.~\ref{eq:surrogate}), the surrogate gap in Eq.~\ref{eq:early_bound} depends on future tokens ($\alpha^{(i)}, i \ge 2$), and is expected to have a higher variance and thus is a worse surrogate loss.
Intuitively, $\mathcal{L}_{late}$ is a better choice for late-stage training because it is simulating the exact SD autoregressive inference, where we essentially avoid parallel predictions and only predict the immediate next token.

However, $\mathcal{L}_{late}$ is \textit{not a good choice} for early-stage training because it requires ``unrolling'' the data multiple times during training, notably adding training time with almost linear increments.
In contrast, setting single-step $s=1$ and assigning maximum parallel predictions $l=k$ as in our early-stage loss will substantially utilize the Transformer architecture by forwarding multiple tokens and improving sample efficiencies.

\subsection{Training Configurations and Observations}
\label{sect:training}

\begin{figure}[h]
    \centering
    \includegraphics[width=0.7\linewidth]{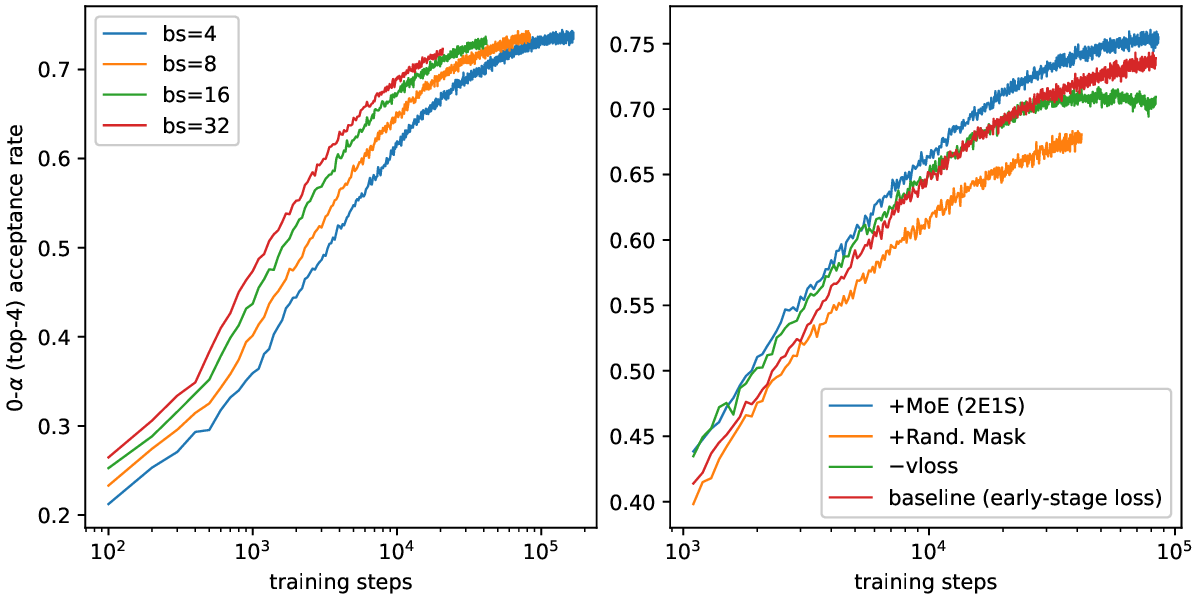}
    \caption{The averaged first-token top-4 acceptance rates using MT-Bench as evaluation set during 10-epoch trainings for the Llama2 7B target model.
    \textbf{Left:} Training with different batch sizes (bs); \textbf{Right:} pilot training experiments using different modeling methods with a fixed bs=8 \textbf{(Right)}.}
    \label{fig:early10ep_training}
\end{figure}
\begin{figure}[h]
    \centering
    \includegraphics[width=1\linewidth]{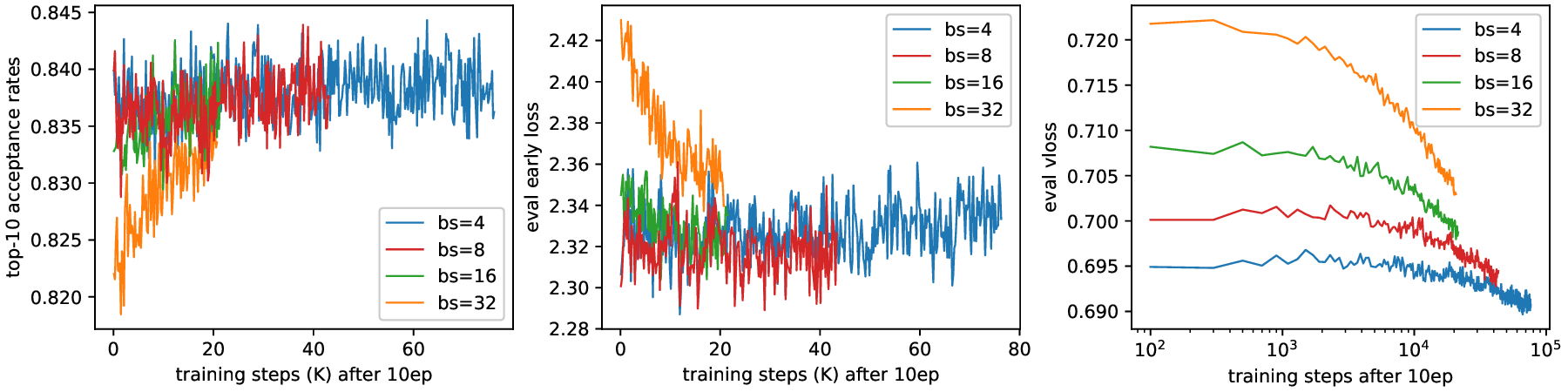}
    \caption{The convergence of training (top-10 averaged acceptance rates, $\mathcal{L}_{early}$, and \textit{vloss} or the regression loss~\parencite{li2025eagle1} after 10 epochs for the Llama2 7B target model. For different batch sizes, the acceptance rates can converge to a similar level.}
    \label{fig:post10ep_training}
\end{figure}

\begin{figure}[h]
    \centering
    \includegraphics[width=1\linewidth]{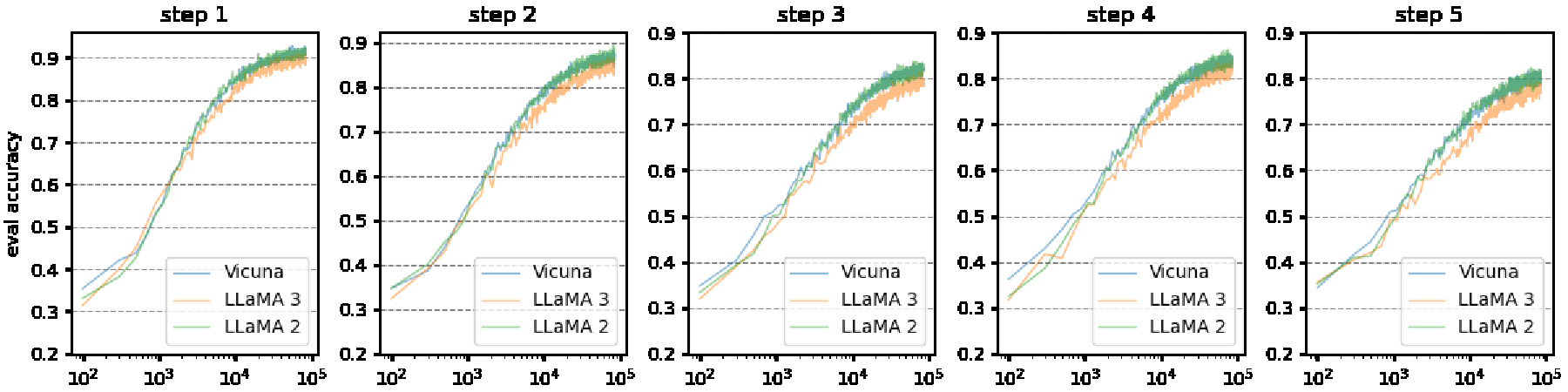}
    \caption{Early-stage (1 -- 10 epochs) training of target models using a window size of $5$.}
    \label{fig:diff_models_early}
    \vspace{0.2in}
    \includegraphics[width=1\linewidth]{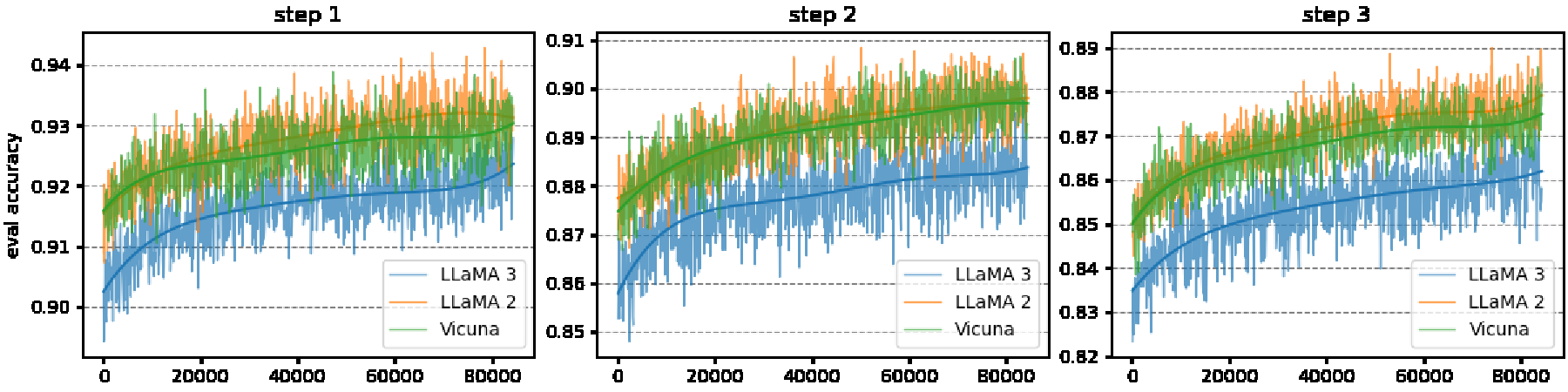}
    \caption{Late-stage training (10 -- 20 epochs) of target models using $3$ simulation steps.}
    \label{fig:diff_models_late}
\end{figure}

We use a context length of 2048 and a training precision of TF32 in different training settings.
We have considered using batch sizes of 4, 8, 16, and 32. Our pivot training in Figure~\ref{fig:early10ep_training} shows the model can converge to a similar level, less sensitive to batch size.
Nevertheless, we choose to use a batch size of 16 for the majority of experiments.
Our optimization uses the PyTorch fused version of AdamW~\parencite{loshchilov2019adamW} kernel with betas (0.9, 0.95), and we use a constant learning rate of 3e-5 with a warm-up of 2000 steps.
In addition, we adopt a maximum gradient norm of 0.5.

To align our training when replicating EAGLE~\parencite{li2025eagle1}, we enforce the same training settings in EAGLE in addition to aforementioned settings.
This also includes adding a Gaussian noise $N(0, 0.2)$ to target model states, and importantly, adding a hidden state distillation loss (i.e., \textit{vloss} or regression loss) with a coefficient of 10 in both stages to regularize the training.
In our pilot training experiments (Figure~\ref{fig:early10ep_training}, right), without adding this distillation loss will cause the model to even degrade at the end of the training.

Specific to our modeling, we set the early-stage window $k = 5$, and due to time and expense budgets, we limit the simulation steps $s=4$ unless described otherwise.
For $s = 3$, a single GPU with 24GiB memory is sufficient to train a 7B model using an unit batch size, although we choose to use the A6000 Ada to train our most competitive models with $s = 4$ and a larger batch size of $16$ in most of experiments (according to Figure~\ref{fig:ablation_methods} and our pivot trainings in Figure~\ref{fig:post10ep_training}, the differences of the final model should be minimal).
After 10 epochs, we find that our early-stage loss starts to converge (Figure~\ref{fig:post10ep_training}), although the incorporated EAGLE vloss can still improve.

Our final model training processes for both stages can be found in Figure~\ref{fig:diff_models_early} and \ref{fig:diff_models_late}.

\subsection{Ablation Study}

\begin{figure}[h]
\begin{minipage}{0.3\textwidth}
\hfill%
\includegraphics[width=0.89\linewidth]{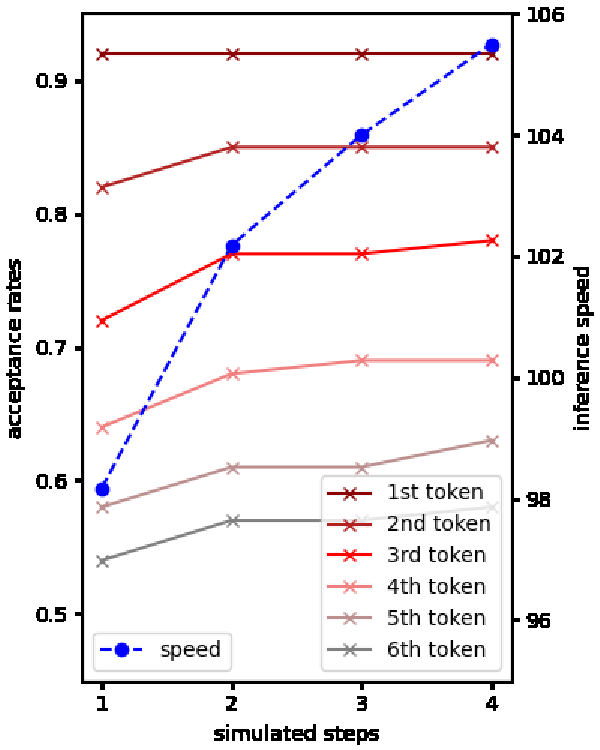}
\vspace{-0.05in}
\end{minipage}%
\begin{minipage}{0.7\textwidth}
\includegraphics[width=\linewidth]{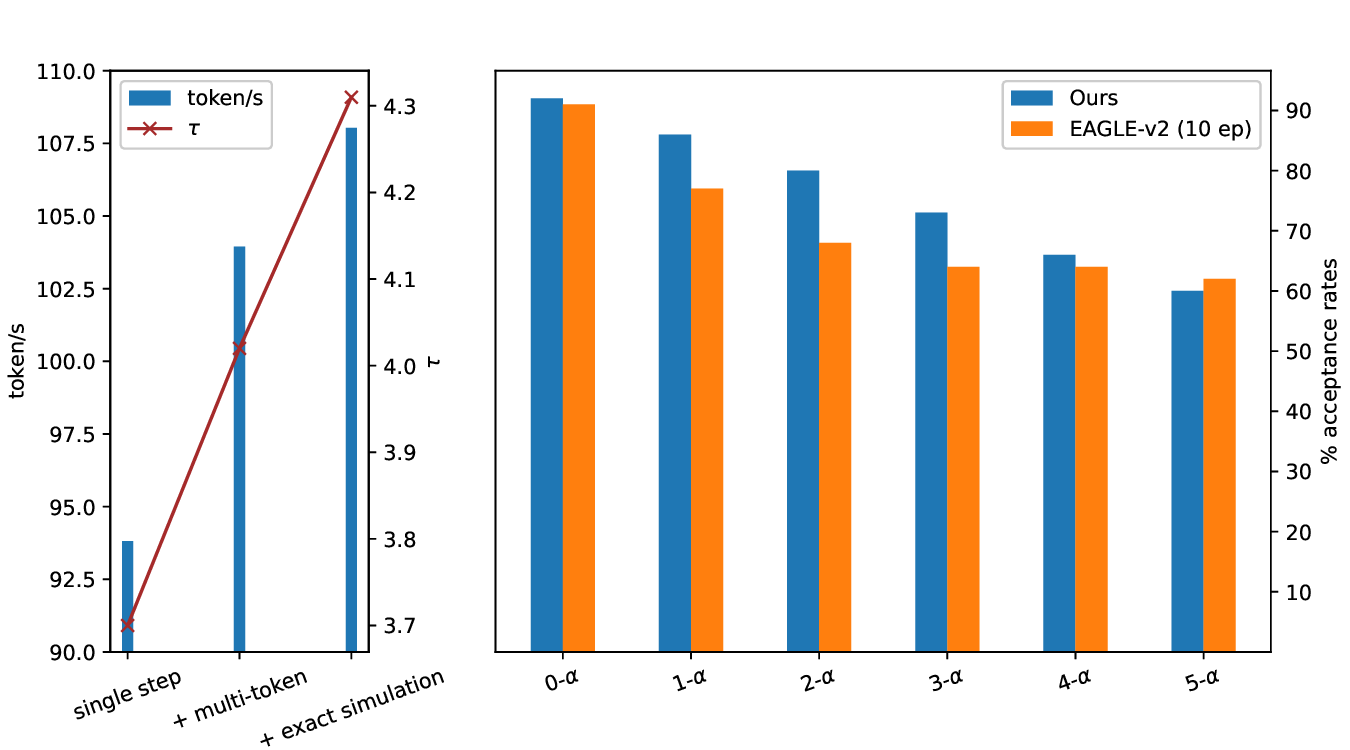}
\end{minipage}
\caption{Ablations of our methods evaluated on MT-Bench. \textbf{Left:} Late-stage training effectiveness at different token positions using different training settings; \textbf{Middle:} The end speed (in token/s) and acceptance length ablations for the major methods proposed in this work; \textbf{Right:} Our model after two-stage training compared to EAGLE-v2 trained for only the early stage.}
\label{fig:ablation_methods}
\end{figure}

\begin{figure}[h]
    \centering
    \includegraphics[width=.8\linewidth]{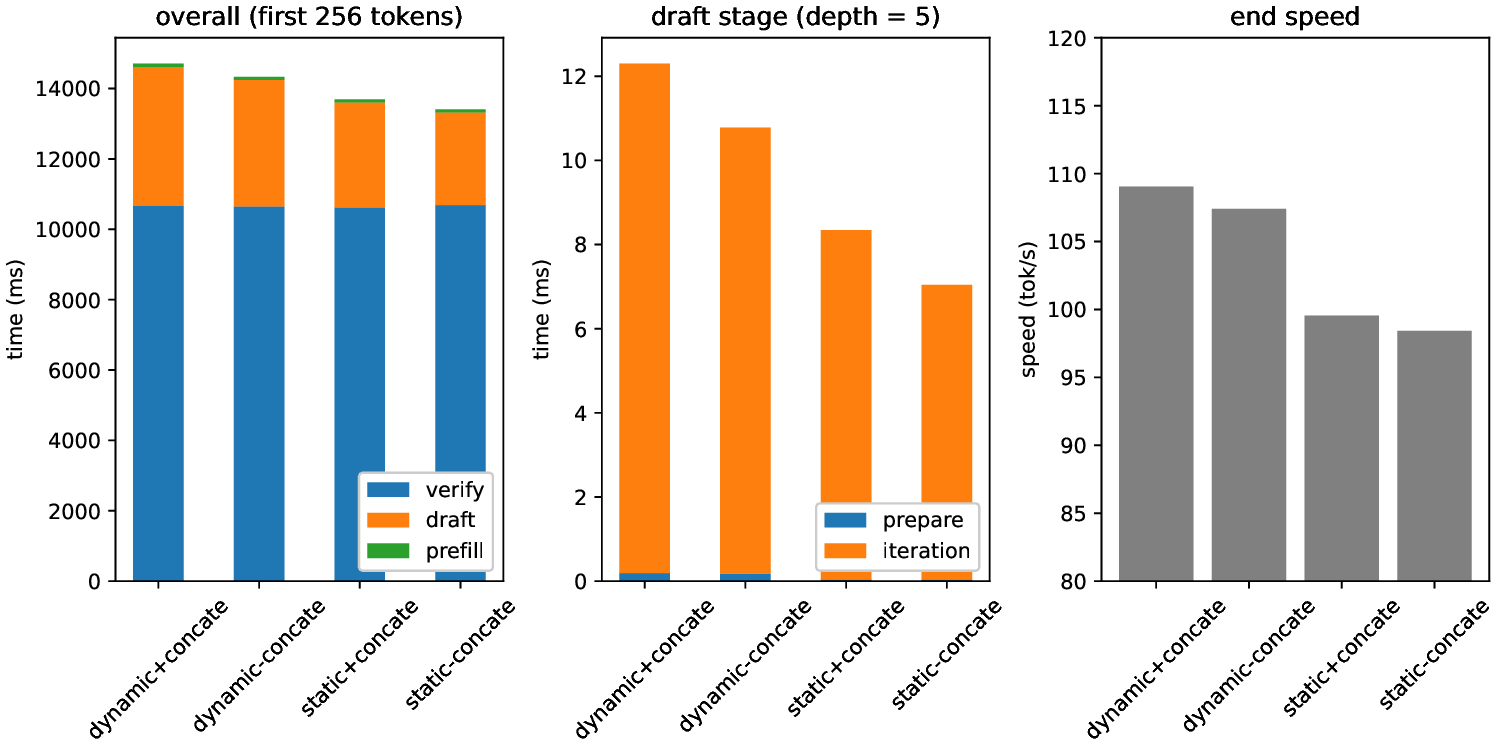}
    \caption{The decoding cost effectiveness analysis on partial MT-Bench (LLaMA 7B). We compare different settings including whether using dynamic tree attention or concatenating key/value states during the drafting stage in cross-attention heads. \textbf{Left:} overall time cost for first 256 tokens assuming constant compositional costs; \textbf{Middle:} the detailed draft costs; \textbf{Right:} and corresponding final inference speeds.}
    \label{fig:efficiency_breakdown}
\end{figure}

We have done additional studies to break down the improvements in both effectiveness and efficiency.

For effectiveness, we study the impact of simulated steps in late-stage training.
As shown in Figure~\ref{fig:ablation_methods}, both token-wise acceptance rates and inference speeds can be improved consistently by running more simulated steps. 
The further a token is in the prediction window, the more potential it has to benefit from a larger number of simulated steps.
And methodology-wise, the two proposed training stages greatly improve the acceptance length almost linearly, i.e., multi-token prediction has around 0.3 average acceptance length improvements over single-step NTP, and training-time exact simulation further adds a similar improvement after the late-stage training.

Moreover, our effectiveness improvements in the late stage, compared to EAGLE-v2, mainly comes from mid-range tokens (2nd to 3rd tokens) where we maintain notably higher numbers in terms of per-token acceptance rates (Figure~\ref{fig:ablation_methods}, right). 
This may be attributed to our cross-attention architecture and the condensing of future token information in the early stage of the training.

In terms of efficiency, we provide a decomposition of the time costs for the overall generation process, projected over the first 256 tokens, as well as the per-iteration cost of proposing draft tokens within a single SD iteration (Figure~\ref{fig:efficiency_breakdown}).
Our different configurations may contribute to runtime costs differently.
Particularly, we denote our method \textit{dynamic} if we apply a dynamic drafting attention tree similar to EAGLE, and we denote our method \textit{concate} when we do concatenation with newly generated hidden stages to be used for generating the next token.

As shown in Figure~\ref{fig:efficiency_breakdown}, although our optimal configuration is ``dynamic+concate'', we find that concatenation can be disabled with only up to 3\% loss in speed.
This option is potentially valuable for deployment on devices with high memory movement penalties, where the efficiency gains from avoiding the copying and concatenation of dynamically generated states may outweigh the accuracy degradation.
However, achieving such flexibility is more challenging in self-attention heads where autoregressively generated states must be appended to predict subsequent tokens.

However, we observe a substantial speed loss when drafting tokens using static trees compared to dynamic trees (Figure~\ref{fig:efficiency_breakdown}, right).
This further underscores the importance of prediction accuracy in the speculative decoding trade-off, as static trees incur greater losses in accuracy than the gains they offer in iteration efficiency.

\subsection{Supplementary Explorations}

\begin{figure}
    \centering
    \includegraphics[width=1\linewidth]{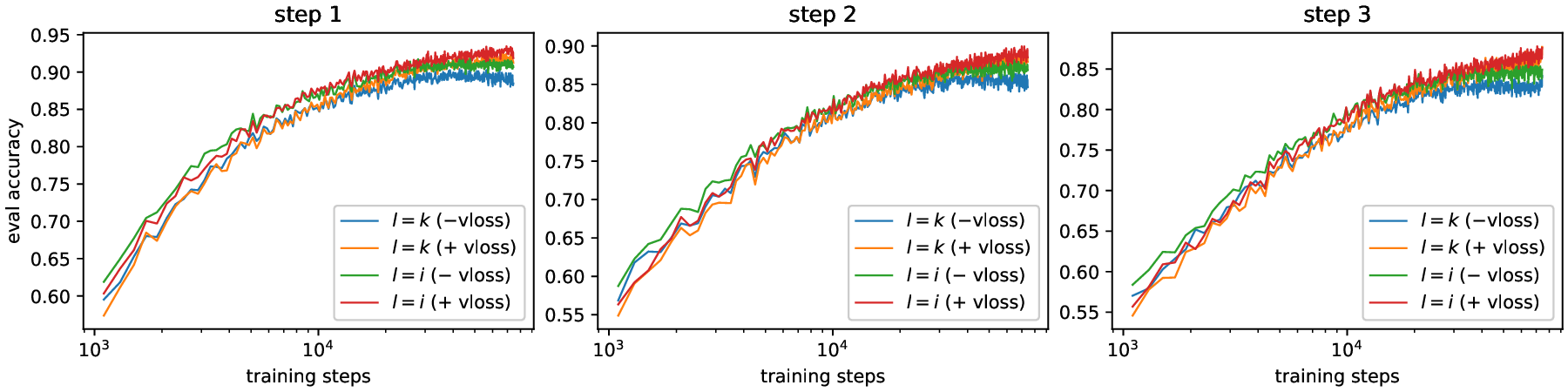}
    \caption{Various loss alternatives (trained from scratch with a window of $3$ and simulated for all $3$ steps at each training iteration).}
    \label{fig:misuse_losses}
\end{figure}

During the pilot training, we have also explored adding random masks and replacing the MLP layer with MoE~\parencite{dai2024deepseekmoe}\footnote{Due to GPU memory constraints, we have reduced the default routed experts to the minimal: only 2 routed experts and 1 shared experts.} as shown in Figure~\ref{fig:post10ep_training}.
However, random masks lead to underperforming our proposed early-stage training, and although MoE improves acceptance rates, its added overheads make the resulting draft model less efficient.
Nevertheless, we believe a specialized MoE kernel that optimizes inference speed may greatly reduce these overheads, but we leave this to future work.

In addition, we have tried various general loss forms in Eq.~\ref {eq:late_loss}, but trained for the initial 10 epochs. As shown in Figure~\ref{fig:misuse_losses}, we can verify that the proposed exact simulation (when $l=i$) achieves better evaluation accuracies compared to alternative combinations.
The $l=k$ combines both multi-token prediction with multi-step simulations, but causes suboptimal accuracies for all 3 steps.
Lastly, we find that the regularization loss (vloss) is crucial to the converged performance in both cases.

\end{document}